# A Broad-Coverage Deep Semantic Lexicon for Verbs

James Allen[1,2], Hannah An[1], Ritwik Bose[1], Will de Beaumont[2] & Choh Man Teng[2]

[1]University of Rochester, [2]Institute of Human and Machine Cognition

[1]Dept. Computer Science, Rochester, NY 14627 USA, [2]40 S. Alcaniz, Pensacola, FL 32501 USA

{jallen,wbeaumont,cmteng}@ihmc.us, {yan2,rbose}@cs.rochester.edu

**Abstract**

Progress on deep language understanding is inhibited by the lack of a broad coverage lexicon that connects linguistic behavior to ontological concepts and axioms. We have developed COLLIE-V, a deep lexical resource for verbs, with the coverage of WordNet and syntactic and semantic details that meet or exceed existing resources. Bootstrapping from a hand-built lexicon and ontology, new ontological concepts and lexical entries, together with semantic role preferences and entailment axioms, are automatically derived by combining multiple constraints from parsing dictionary definitions and examples. We evaluated the accuracy of the technique along a number of different dimensions and were able to obtain high accuracy in deriving new concepts and lexical entries. COLLIE-V is publicly available.

## 1. Introduction

While there are significant ongoing efforts to develop deeper language understanding systems, progress is hampered by the lack of a broad-coverage deep semantic lexicon linked to ontologies that can support reasoning about commonsense knowledge. By *broad-coverage* we mean that the lexicon substantially covers typical English usage (on the scale of WordNet (Fellbaum, 1998)). By *deep* we mean that all words are assigned senses that are organized into an ontology, and that each sense has associated semantic roles with semantic preferences and syntactic linking templates. By ontology we mean not only a hierarchy of concepts with inheritance of properties, but also axioms that capture the relationships between concepts, especially temporal and causal relationships for events.

We have developed *Comprehensive OntoLogy and Lexicon In English* (*COLLIE*), a new resource constructed by bootstrapping from an initial hand-built lexicon and ontology, and using a semantic parser to read definitions automatically from dictionaries to extend the coverage. Here we report on the verb component (*COLLIE-V*) which is the first part of the overall effort to build an extensive semantic lexicon. The full COLLIE, including nominal concepts and properties, will be released at a later date.

As an example showing the typical contents of COLLIE-V, consider the entry for one sense of the verb *kill,* which we will call ONT::KILL. The information about this sense is captured in two places: (1) information in the ontology about the concept ONT::KILL (which contains other verbs such as *murder* and *slaughter* in addition to *kill*), and (2) information in the lexicon about the behavior of the verb *kill*.

Looking first at the concept, ONT::KILL is a subconcept of ONT::DESTROY, which involves events of destruction. ONT::KILL has two core semantic roles, AGENT (playing a causal role in the event) and AFFECTED (the object affected by the event). Furthermore, the roles are associated with semantic preferences. For example, the AFFECTED role of ONT::KILL has a preference to apply to entities that can be alive (another concept in the ontology).

Also associated with ONT::KILL is an axiom that says *if X kills Y, then X causes Y to die.* This involves another concept in the ontology, ONT::DIE, with its own axiom that says it entails a transition from being alive to being dead. (And further, the concepts of alive and dead are also defined in the ontology).

The lexical item *kill*, on the other hand, is associated with linking templates that relate syntactic realizations to the semantic roles of ONT::KILL (cf. VerbNet frames (Kipper et al., 2008)). For *kill*, one template indicates the transitive use: the subject fills the AGENT role and the direct object fills the AFFECTED role. Expressed in VerbNet notation, the linking template is

AGENT **V** AFFECTED[+LIVING]

*Kill* also has another template for an intransitive form involving just the AGENT role to handle generic statements such as *Pesticides kill*.

Table 1 summarizes many of the currently available lexical resources, none of which meet all the requirements for a comprehensive resource. WordNet has the most extensive coverage, with over 13,000 verb senses with associated definitions (as natural language glosses) and examples. Verbs in WordNet is organized into a hypernym hierarchy which is often used as an ontology surrogate. But this hierarchy is highly incomplete. Specifically, about 550 verbs have no hypernym (supertype). Still, because of its coverage, WordNet is the most widely used resource for semantic information.

VerbNet provides extensive information on semantic roles, linking templates, and causal-temporal axioms for broad classes of verbs. However, VerbNet organizes verbs into only 329 verb classes. Because each class covers a wide range of diverse verbs, VerbNet does not provide a fine-grained semantics and the axioms defined for each class are necessarily quite abstract. For example, *abate, gray, hybridize* and *reverse* are all in the same class and share the same defining axiom. Still, VerbNet represents the state of the art in providing axioms related to verbs.

Propbank (Palmer et al., 2005) is commonly used as a taxonomy of verb senses, and is used in the Abstract Meaning Representation (AMR) (Banarescu et al., 2013). Each entry identifies a set of semantic roles related more to syntax than semantics. While Propbank identifies senses for individual verbs, it does not group similar verbs that would identify closely related concepts or categorize the senses into an ontology. For instance, PropBank indicates no relation between similar verbs such as *contradict*, *refute* and *disagree* and no connections between *kill* and *die*.

| Lexical Resource | Coverage (Verbs; Sense Types; Avg # Senses/Verb) | Pros | Cons |
|---|---|---|---|
| WordNet | 11531; 13782; 2.17 | Extensive coverage, hypernym hierarchy, example sentences and templates | Hypernym hierarchy is only a partial ontology, no axioms |
| VerbNet | 4577; 329; 1.48 | Linking templates, semantic roles, selectional restrictions, entailment axioms associated with classes | Limited coverage, only verbs, no ontology, classes and axioms are highly abstract |
| Propbank/ Ontonotes | 2681; 6267; 2.34 | Carefully curated senses, examples, semantic roles, informal linking templates | Limited coverage, mostly verbs, no axioms or ontology |
| Framenet | 3349; 1086; 1.56 | Organized by conceptual frames, semantic roles, linking templates implicit from examples | Limited coverage, idiosyncratic semantic roles, no axioms, limited ontology |
| TRIPS | 2567; 980; 1.21 | Linking templates, semantic roles, selectional preferences, fully integrated into ontology | Limited coverage, no axioms |
| COLLIE-V | 9654; 6516; 2.30 | Linking templates, semantic roles, selectional preferences, fully integrated into ontology, axioms, extensive coverage | Some errors due to semi-automatic nature of construction |

Table 1. A comparison of existing lexical resources

OntoNotes (Weischedel et al., 2011) provides a sense inventory by grouping WordNet senses with high inter-annotator agreement. PropBank frames are linked to OntoNotes senses. Some of the OntoNotes senses are further clustered into the Omega 5 ontology. However, many of the verb senses and most of the PropBank frames are not mapped to the ontology. For instance, Omega 5 contains entries for *disagree*, but not *contradict* or *refute*. Similarly it contains *kill* but not *die*.

Framenet (Baker et al., 1998) organizes its lexicon around conceptual frames capturing everyday knowledge of events. While it provides informal definitions of these classes, there are no axioms. In addition, unlike the other resources, the semantic roles in Framenet are idiosyncratic to each class.

TRIPS provides detailed lexical information integrated with an ontology. It uses a principled set of semantic roles (Allen & Teng, 2018) with selectional preferences expressed in the ontology, as well as linking templates. The main weaknesses are the coverage and the lack of axioms. A significant advantage of TRIPS is that it is connected to a publicly available domain-independent, broad-coverage semantic parser (Allen & Teng, 2017).

In summary, none of these existing resources come close to being the broad-coverage deep semantic lexicon we seek. While WordNet has extensive coverage it lacks the details available in other resources such as VerbNet. All the other resources considered fall short on coverage, integration with an ontology (except TRIPS), and effective entailment axioms (except VerbNet).

There has also been significant interest in acquiring knowledge using information extraction techniques and organizing such knowledge based on distributional semantics (e.g., Etzioni et al, 2011; Carlson et al, 2010). Such work, however, remains close to the surface level of language, involving mostly uninterpreted words and phrases and surface relations between them (e.g., is-a-subject-of, is-an-object-of), or a limited number of pre-specified relations organized in a fairly flat hierarchy. In addition, information extraction tends to focus more on learning facts (e.g., Rome is the capital of Italy) rather than conceptual knowledge (e.g., kill means cause to die). This work does not address the central goal of this paper, namely building a rich semantic lexicon and associated ontology.

As just mentioned, this paper describes a comprehensive verb resource linked to an event ontology. We describe how the resource was built semi-automatically, by bootstrapping from hand-built existing resources, applying multiple consistency and validity tests, with some human post correction and reprocessing to eliminate errors. The resource is publicly available in several different formats.

## 2. The Bootstrapping System

The core idea is to start with an initial lexicon and ontology, and iteratively expand to full coverage by parsing word sense definitions. We chose the TRIPS lexicon and ontology as the starting point partly because of the availability of the fairly robust and accurate TRIPS parser that can map definitions into detailed logical forms expressed in the TRIPS ontology. Note that the existing TRIPS ontology and lexicon cover English in all parts-of-speech, thus providing a uniform representation when parsing the definitions and also a good foundation for building up COLLIE-V.

In addition, TRIPS contains a mapping between its ontology types and WordNet synsets which allows the TRIPS parser to produce semantic representations of sentences even if they involve words not explicitly in the TRIPS lexicon. Figure 2 shows a fragment of the WN-TRIPS mapping. We see a direct mapping from the WN synset *have%2:34:00::* to the TRIPS concept ONT::CONSUME. The synsets *eat%2:34:01::* and

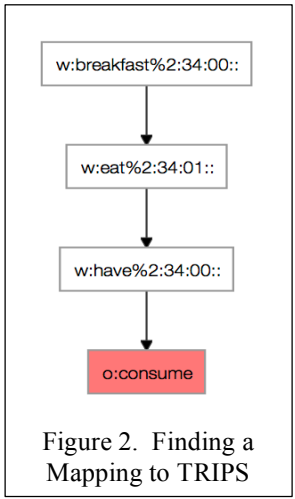

Figure 2. Finding a Mapping to TRIPS

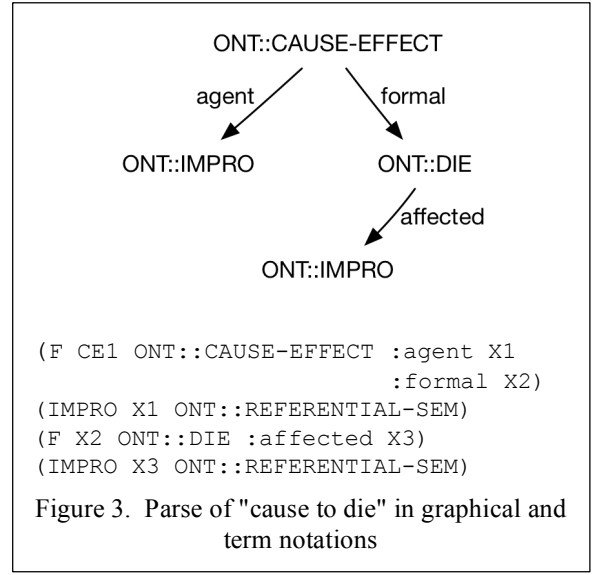

```
(F CE1 ONT::CAUSE-EFFECT :agent X1
                        :formal X2)
(IMPRO X1 ONT::REFERENTIAL-SEM)
(F X2 ONT::DIE :affected X3)
(IMPRO X3 ONT::REFERENTIAL-SEM)
```

Figure 3. Parse of "cause to die" in graphical and term notations

*breakfast%2:34:00::*, one the other hand, have indirect mappings to ONT::CONSUME via the WordNet hypernym links. Any words associated with these synsets that are not explicitly in the TRIPS lexicon are interpreted as having a sense ONT::CONSUME.

More details of the TRIPS parser and ontology, including the logical form representation and parsing framework, can be found in Allen & Teng (2017, 2018).

### 2.1 Extending TRIPS-WordNet Mappings

Essential to the success of this approach is a high-quality mapping of WordNet synsets into the TRIPS ontology. So the first step in this project was to clean up and extend the TRIPS-WordNet mappings. Specifically, we

- detected duplicate mappings and either removed or remapped them;
- created mappings for synsets that have no induced mappings (i.e., there are no mapping for the synset or its inherited hypernyms);
- identified and remapped synsets that are mapped to broad abstract concepts in the TRIPS ontology;
- linked verb event synsets with their nominalizations (which are not hierarchically related in WordNet)

Note that as a result of this effort we have created a resource that is useful in its own right independent of our main goal. Specifically, we have created an upper ontology for WordNet that fills in the gaps in the WordNet hypernym hierarchies. As one of many examples, WordNet has a sense of *insist* that has no hypernym, and so appears semantically unrelated to other verbs, including very similar ones such as *demand*. Furthermore, the verb *insist* is not related specifically to the nominal concept *insistence* (other than the fairly broad relation "derivationally related form"). In the expanded ontology, these senses are all closely related. We will report on the evaluation of the expanded ontology with respect to human similarity judgements in Section 4. This ontology is also available as described in Section 6.

### 2.2 Entailment Axioms

A key contribution of this paper is generating entailment axioms for the types in the ontology, both for existing types and newly learned types. For readability, we will represent axioms in an abbreviated form in which the semantic role names for arguments are suppressed. For example, an entailment axiom associated with the concept ONT::KILL is written as follows.

```
[ONT::KILL ?agent ?affected] =>
    [ONT::CAUSE-EFFECT ?agent
        [ONT::DIE ?affected]]
```

The `?agent` and `?affected` terms are universal variables denoting the AGENT and AFFECTED roles of the ONT::KILL concept. Since the left hand side (the antecedent) of a rule is uniquely determined by the concept and its semantic roles, we will generally only show the right hand side (the consequence) in subsequent examples. Note also that other important roles, such as the temporal role indicating the time of the event, have also been omitted for simplicity. For example, the entailed ONT::CAUSE-EFFECT event co-occurs with the ONT::KILL event. In other words, whenever a *killing* occurs, at that time the killer *causes* the killee to die.

While such axioms seem simple, by chaining sets of these axioms more complex entailments can arise. For instance, combining the definition for *kill* above with a definition of *die*:

```
[ONT::BECOME ?affected
    [ONT::DEAD ?affected]]
```

and a definition of *become*, we can then infer that if X kills Y at time T, then Y is not dead at T, but is dead immediately after T. Thus, by simple chaining, we can derive a wide range of consequences.

### 2.3 Grammar for Parsing Definitions

We generated entailment axioms by automatically reading WordNet glosses. To improve word sense disambiguation during parsing, we used the Princeton WordNet Gloss Corpus, which has some words in the definitions tagged with their WordNet sense tags. For example, one of the mappings to ONT::KILL is the WordNet sense

```
kill%2:35:00:: "cause to die%2:30:00::".
```

The TRIPS parser already is able to produce grammatical analyses containing gaps or traces in order to handle questions, relative clauses, and other phenomena. To convert the parser to reading definitions, we added about a dozen top-level grammar rules that accept various constituents with gaps, represented as implicit terms called IMPROs in the TRIPS logical form. With this extension, the definition "cause to die" is parsed to the logical form shown in Figure 3 (in both graphical notation and term notation), slightly abbreviated and simplified for readability. Here, ONT::REFERENTIAL-SEM is a high level type in the TRIPS ontology that includes all objects that can be a referent.

As we will show in more detail later, the gaps in the definitions are strong signals for the roles of the predicate being defined. In this case, the first IMPRO (X1) corresponds to the AGENT role of ONT::KILL, and the second (X3) corresponds to the AFFECTED role. It is a relatively simple transformation to take such a definition and create an entailment axiom as shown earlier.

Note that this is one of the simplest examples. In general, parsing WordNet definitions poses many challenges. The definitions can be arbitrarily complex, with many conjunctions and disjunctions and attendant problems of attachment. Gaps (IMPROs) do not contain any semantic information, which hampers sense disambiguation as TRIPS relies on semantic preferences to identify the correct senses that can fill a semantic role. Furthermore, it is often possible to construct a parse for both the transitive (two IMPROs) and intransitive (one IMPRO) uses of a verb, especially in the absence of semantic guidance from the IMPROs.

## 3. Learning New Concepts

To construct a new concept and the lexical items that realize it, we need to learn all of the aspects discussed above, including identifying the location of the new concept in the ontology, the core semantic roles, entailment axioms, semantic preferences (cf. selectional restrictions) for the roles and, for each of the lexical realizations of the concept, the linking templates that map syntactic arguments to semantic arguments. New concepts are placed in the ontology by combining evidence from the WN-TRIPS mappings and using classification algorithm on the definitions as in Description Logics (Baader et al., 2007).

### 3.1 Initial Information from the Mappings

While the centerpiece of this work is using the synset definitions in WordNet to derive new lexical items and concepts, we should first observe that this process is not performed in isolation. We already have a semantic characterization of any synset that has a (direct or indirect) WordNet-TRIPS mapping. For example, consider the synset *breakfast%2:34:00::* (eat an early morning meal), with the mappings shown in Figure 2. Although *breakfast%2:34:00::* does not have a direct mapping to TRIPS, by searching up the WordNet hypernym hierarchy, we can infer that *breakfast%2:34:00::* is a subclass of ONT::CONSUME since there is a mapping from the hypernym *have%2:34:00::* to ONT::CONSUME. Inheriting the role and semantic preference specifications of ONT::CONSUME, we also know *breakfast%2:34:00::* might involve the semantic roles AGENT, AFFECTED and RESULT. Furthermore, we know that the AGENT is typically a living entity, and that the AFFECTED is a tangible, comestible object. Finally, looking at other verbs that are related to ONT::CONSUME, we can identify possible linking templates, including the AGENT Template (simple intransitive) and the AGENT-AFFECTED Template (simple transitive).

All this information is used to constrain and guide the processing of the definition of *breakfast%2:34:00::*, described in the next sections. Most important, these constraints allow us to flag cases where the definition is likely unusable, either because of parsing errors or because it is a poor definition of the concept.

### 3.2 Identifying the Semantic Roles

As discussed above, there is a relationship between the semantic roles of a target word sense and those used in its definition. One of the most common cases is that the elided roles in the definition correspond to the unfilled roles in the target. For instance, from the definition < > *cause* < > *to die*, we can lift the two elided roles to the target word, i.e., < > *kill* < >. Allen and Teng (2018) reported good accuracy in using gaps to identify semantic roles. We adopted and expanded on this approach.

We identified a number of ways such elided roles (IMPROs) can be realized in a parse. For instance, for *outweigh*, the IMPROs occur in an embedded clause in the parse of its definition *be heavier than*. Some of these patterns and corresponding parse skeletons are shown in Table 4 and Figure 5. The rules for identifying the roles are ordered such that where multiple rules are applicable, only the one with the highest priority would be selected.

We also identified other patterns commonly used for definitions. These include the use of indefinite pronouns (e.g., bring: take *something or somebody* with oneself somewhere); enclosing an argument in parentheses (e.g., preserve: prevent (*food*) from rotting); the use of indefinite nouns (e.g., drink: take in *liquids*); and specific words such as "certain" and "particular" (e.g., charge: set or ask for *a certain price*).

| Parse Skeleton | Role for $\phi_{IMPRO}$ | Example ($\phi$ denotes target argument) | Rationale |
|---|---|---|---|
| $V_1 \xrightarrow{CORE\text{-}ROLE} \phi_{IMPRO}$ | CORE-ROLE | censure: $\phi$ rebuke $\psi$ formally | The default scenario where the IMPRO is lifted to the new event |
| $V_1 \xrightarrow{FORMAL} V_2 \xrightarrow{CORE\text{-}ROLE} \phi_{IMPRO}$ | AFFECTED | agitate: $\psi$ cause $\phi$ to be excited | The embedded IMPRO is affected by the new event via CORE-ROLE |
| $V_1 \xrightarrow{CORE\text{-}ROLE} P_2 \xrightarrow{FIGURE} \phi_{IMPRO}$ | CORE-ROLE | weaken: $\psi$ lessen the strength of $\phi$ | The FIGURE role is lifted to substitute for $P_2$ |
| $V_1 \xrightarrow{RESULT} P_2 \xrightarrow{GROUND} \phi_{IMPRO}$ | NEUTRAL | approach: $\psi$ move towards $\phi$ | The GROUND is unchanged by the new event |
| $V_1 \xrightarrow{FORMAL} P_2 \xrightarrow{COMPAR} \phi_{IMPRO}$ | NEUTRAL1 | outweigh: $\psi$ be heavier than $\phi$ | The entity being compared to is unchanged by the new event |

Table 4. Rules for identifying semantic roles from parsing definitions. CORE-ROLE refers to any of the main roles (and variants) in TRIPS: AGENT, AFFECTED, NEUTRAL, EXPERIENCER and FORMAL.

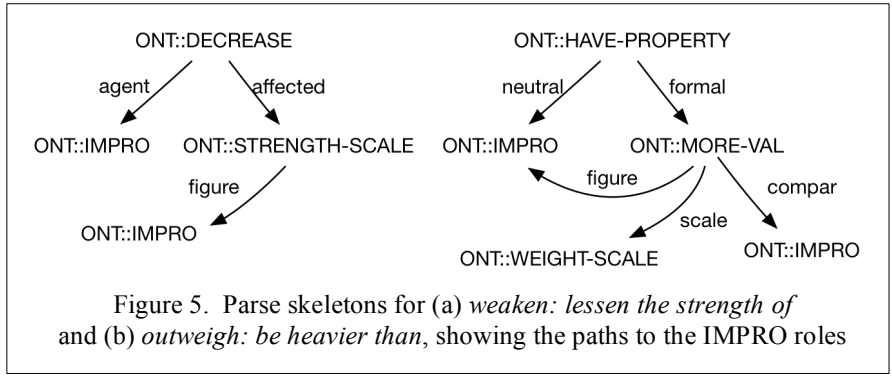

Figure 5. Parse skeletons for (a) *weaken: lessen the strength of* and (b) *outweigh: be heavier than*, showing the paths to the IMPRO roles

### 3.3 Identifying Semantic Preferences

We can also obtain semantic preferences on the roles from their places in the definitions. For example, from the definition *kill: cause to die*, we can reason that since the AFFECTED role of *kill* is identified with the AFFECTED role of *die*, the semantic preferences of the latter (e.g., that it is a living entity) should transfer over as well.

One might notice that some of the patterns for identifying roles also give information about the types of objects that can be used to instantiate these roles. For example, by parsing the definition of *remit: send (money) in payment*, we can derive that *remit* takes an AFFECTED role, and should be filled by the ontology type to which *money* belongs.

### 3.4 Identifying Linking Templates

We use several different strategies for identifying possible linking templates. First, we exploit the property, first noted by Levin (1993), that words that allow similar sets of alternations often are semantically similar. We use this observation in reverse. Given a new word and type, we look at semantically similar words (that is, the words that belong to the same ontology type) and hypothesize the new word behaves in a similar way and thus can use the same templates as those used by these other words. This procedure often over-generates the pool of templates, so we then filter them based on compatibility with the semantic roles that were identified from the definition. In other words, proposed templates that would require a role not identified from the definition would be discarded. For example, a sense of *incite* in WordNet is classified under ONT::CAUSE-EFFECT. One of the words associated with ONT::CAUSE-EFFECT is *force* which has a template AGENT-FORMAL-SUBJCONTROL-TEMPL. This requires a FORMAL role which was not derived from the definition of *incite*. Thus this template is removed as a candidate. The templates for *incite* are the ones that involve only the AGENT and AFFECTED roles. Finally, we use a backoff strategy if suitable linking templates have not been identified, and introduce templates that are most commonly associated with the set of semantic roles that were identified for the new type, for instance the transitive template with AGENT and AFFECTED roles.

### 3.5 Classification into the Ontology

We noted in Section 3.1 that by using the TRIPS-WordNet mapping, *breakfast%2:34:00::* can be preliminarily placed in a new ontology type directly under ONT::CONSUME. After parsing its definition *eat an early morning meal*, we further infer that it could be placed directly under ONT::EAT since breakfasting is defined in terms of eating (and the verb *eat* belongs to the type ONT::EAT). Noting that ONT::EAT is a subtype of ONT::CONSUME in the TRIPS ontology (so everything is consistent) we place the type for *breakfast%2:34:00::* under the more specific ONT::EAT instead of ONT::CONSUME.

When incorporating a new type derived from a WordNet synset into the TRIPS ontology, in addition to considering subsumption relations between the main predicate of the WordNet definition and the TRIPS ontology types (as above), we also extend the comparison to their semantic roles. For example, by following the hypernym hierarchy until a WordNet-TRIPS mapping is found, *port%2:34:00::*, *claret%2:34:00::* and *wine%2:34:00::*, defined as *drink port, drink claret* and *drink wine* respectively, are all preliminarily placed under ONT::CONSUME. Since the main predicates of all three definitions are the verb *drink*, we further infer that these synsets could be under the more specific TRIPS ontology type ONT::DRINKING (which contains the verb *drink*).

So far, this processing is the same as in the breakfast example above. Then, comparing the semantic types of the arguments, we find that *port*, *claret*, and *wine* are all mapped to ONT::ALCOHOL as an AFFECTED role of ONT::DRINKING. This allows us to group *port%2:34:00::, claret%2:34:00::* and *wine%2:34:00::* into a new type under ONT::DRINKING, with an AFFECTED role that has semantic preferences for ONT::ALCOHOL.

We also created a number of hand-mapped rules for processing some common phrasings. For instance, ONT::MOVE-RAPIDLY is characterized by the main predicate ONT::MOVE with a modifier ONT::SPEEDY. When the definition of a WordNet synset matches one of these rules, we place the synset in a new type under the type of the matched rule. For example, the definition of *breeze%2:38:00::* (to proceed quickly and easily) is parsed as ONT::GO-ON (proceed) with the modifiers ONT::SPEEDY (quickly) and ONT::EASY (easily). By regular processing we infer that *breeze%2:38:00::* could be placed under ONT::GO-ON. However, since ONT::GO-ON is a subtype of ONT::MOVE and the definition also contains the modifier ONT::SPEEDY, the definition matches the rule for ONT::MOVE-RAPIDLY and we place the type for *breeze%2:38:00::* under ONT::MOVE-RAPIDLY instead of ONT::GO-ON. These rules allow us to place more weight on auxiliary parts of the parse, in this case the manner of moving.

## 3.6 Consistency Checking and Manual Correction

The system automatically conducts consistency checking on the preliminary results so that if the ontology type, semantic roles and semantic preferences are not consistent with the constraints obtained from the mappings as described in Section 3.1, a new entry would not be created from the definition. These cases were flagged and we manually examined a number of them. We identified a few common problems, some of which could be easily corrected. For example, the WordNet-TRIPS mapping may be in error or the sense tags in the glosses from the Princeton Gloss Corpus may be incorrect. Others are more difficult, e.g., the WordNet definition is poorly worded and does not capture the intended sense, or the definition is too complex and cannot be parsed correctly.

For COLLIE, 12.9% of the verb definitions were automatically rejected based on consistency checking. We believe that strong rejection strategies are essential for building a high quality semantic lexicon. Even though these problematic cases are often just skipped automatically when they fail the consistency checks, many WordNet synsets have multiple definitions, so even if some of the definitions are not usable, we may still use the remaining consistent definitions to generate an entry. In addition, as we expand the ontology and lexicon of the resource using the definitions we could process successfully, we expect some of these problematic definitions would become parsable in subsequent iterations of the processing when based on a more extensive resource.

## 4. Evaluation

Summary statistics of COLLIE-V are given in Table 1. Of the 13782 WordNet synsets, 12007 were successfully incorporated, creating 5536 new ontology types in COLLIE. This indicates that our approach is not merely replicating one-to-one the WordNet structure (which is often criticized for being too fine-grained) and is fairly successful in organizing entries with similar definitions into a single ontology type.

We further evaluated COLLIE along the following dimensions.

### 4.1 The Ontology

As discussed above, our system makes use of WordNet and the TRIPS ontology in parallel, using the TRIPS ontology as an upper ontology and augmenting it with the WordNet hypernym hierarchy via mappings between the two. For evaluation we used two word-similarity datasets: (i) SimLex-999 (Hill et al., 2014) consists of 666 noun pairs, 222 verb pairs, and 111 adjective pairs; (ii) SimVerb-3500 (Gerz et al., 2016) consists of 3500 verb pairs. Both datasets focus specifically on similarity rather than relatedness. For example, while coffee and cup are highly related, the two words represent very different concepts.

We evaluated the similarity between concept pairs using the *WuP* hierarchy similarity measure (Wu & Palmer, 1994). For each pair of words, we consider the highest similarity score over all possible pairs of senses. We chose *WuP* because it relies entirely on the structure of the hierarchy. We computed the *WuP* scores for verbs in WordNet and TRIPS respectively, using the SimVerb test set and the verb pairs in SimLex. Table 6 shows the

|  | WordNet | TRIPS |
|---|---|---|
| SimVerb-Test | 0.42 | 0.497 |
| SimLex-Verbs | 0.37 | 0.522 |

Table 6. Spearman's ρ between WuP scores and human similarity judgement for verbs.

| Word Embedding | | | Ontology-based | | |
|---|---|---|---|---|---|
| Levy & Goldberg | Glove | Swartz et al. | WordNet | TRIPS | Hybrid |
| 0.462 | 0.35 | 0.563 | 0.38 | 0.468 | 0.517 |

Table 7. Spearman's ρ with respect to several word embedding and ontology-based approaches over the SimLex-999 data set

Spearman's correlation measured between each of these scores and the human annotated similarity scores. The similarity scores using the TRIPS ontology agree substantially more with human judgement than the scores computed using WordNet. In order to distinguish between lexical mappings and WordNet mappings in TRIPS, we added 1 to the depth of all WordNet-to-TRIPS mappings.

Table 7 compares the ontology-based measures and several main word-embedding based approaches over the full SimLex-999 data set (Levy & Goldberg (2014), Glove (Pennington et al., 2014) and Swartz et al. (2015)). This is not a clean comparison though, for COLLIE-V (integrated with TRIPS) only concerns verbs, but the reported word-embedding results involve three parts-of-speech (nouns, verbs and adjectives). So we used WuP similarity scores computed using WordNet, TRIPS and a hybrid ontology in which noun pairs were evaluated using WordNet and verb and adjective pairs were evaluated using TRIPS. While Swartz et al. (2015) had the best correlation with human judgement, the ontology-based measures, using solely the structural information in the ontologies, were still fairly competitive.

### 4.2 Ablation Experiments for Semantic Roles and Linking Templates

For WordNet synsets with a direct mapping to a TRIPS ontology type, a gold standard entry can be obtained by examining the manually curated information in the TRIPS ontology and lexicon. For evaluation we performed an ablation study. We deleted the TRIPS lexical entries corresponding to those synsets with direct mappings and tried to recreate these entries by processing their WordNet definitions.

**Semantic Roles** Table 8 reports the precision and recall of the sematic roles identified by our system, compared to a baseline that chose AGENT and AFFECTED roles for each entry. Because of the distribution of verbs in English, this baseline performed surprising well, but our technique demonstrated a superior performance, attaining 0.88 precision and an F1 score of 0.85.

**Linking Templates** For linking templates we used the manually created templates associated with the TRIPS lexical entries as the gold standard. This required us to restrict the test to entries where the WordNet synset and the TRIPS type it maps to share at least one common item in the TRIPS lexicon. These items were removed from TRIPS and we compared the generated lexical templates with the templates of the deleted items. As a baseline, we assigned the transitive template with AGENT-AFFECTED roles to each entry as this is the most common template for verbs.

|  | Precision | Recall | F1 |
|---|---|---|---|
| Definitions | 0.88 | 0.82 | 0.85 |
| Baseline | 0.69 | 0.74 | 0.72 |

Table 8. Accuracy of Automatic Identification of Semantic Rolesets from Definitions

|  | Precision | Recall | F1 |
|---|---|---|---|
| Definitions | 0.62 | 0.96 | 0.76 |
| Baseline | 0.46 | 0.30 | 0.36 |

Table 9. Accuracy of Automatic Identification of Linking Templates

Table 9 reports the accuracy of the templates. Note that our methods are designed to optimize recall, as this would allow sentences with unknown (or ablated) lexical items to be successfully parsed. The recall score of 0.96 indicates we are quite successful at that.

### 4.3 Axioms

To evaluate the accuracy of axioms, we used human expert judgement, since gold standard axioms are not available from TRIPS or other resources. We classified the axioms into several classes: 1) *Correct*: The axiom is an accurate definition of the concept; 2) *Valid*: The axiom is an entailment from the event denoted by the verb being defined but is not a complete definition; and 3) *Wrong*: The axiom is not an entailment of the verb being defined. The last category is further divided into those the system detects and rejects and those that are undetected.

Table 10 reports the evaluation by two human judges of the axioms automatically generated in our test set. 62.4% of the results were judged either correct or valid. Of the remaining errors, the system detected and rejected 16.7% of the cases, leaving 20.9% as axioms that were accepted but erroneous. We are actively working to improve these scores by improving parsing as well as developing better techniques for detecting and rejecting faulty axioms. We plan to release updated versions of COLLIE-V as improvements are made.

## 5. Discussion

Previous attempts to use WordNet as an ontology to support reasoning have mainly focused on nouns, because the noun hypernym hierarchy provides a relatively good subclass hierarchy (e.g., Gangemi et al., 2002). The situation is not the same for verbs however. Verbs in WordNet are not organized into an ontology of event types in terms of major conceptual accomplishments and achievements (cf. Vendler, 1957). In fact, Fellbaum (1998) argues against a top-level verb distinction between events and states. This lack of an upper-level ontology for events creates a significant obstacle to unifying WordNet with ontologies that are built to encode commonsense knowledge and support reasoning.

Most prior work linking WordNet to ontologies has involved producing mappings from the synsets into an upper ontology, without developing the intermediate detail. For instance, SUMO (Niles & Pease, 2001) has a comprehensive mapping from WordNet to its upper ontology, but 670 WordNet verb synsets are mapped to the single SUMO class IntentionalProcess (3 equivalences and 667 subsumptions), including senses as diverse as postdate (establish something as being later relative to something else), average (achieve or reach on average), plug (persist in working hard), diet (follow a regimen or a diet, as for health reasons), curtain off (separate by means of a curtain) and capture (succeed in representing or expressing something intangible). While these links connect WordNet into SUMO, they do not provide significant extra knowledge to enable entailments.

There have been several prior attempts to process WordNet glosses to produce axioms that capture entailments. For the most part, these representations are fairly shallow, and resemble an encoding of the syntactic information in a semi-formal logical notation, with each word represented as a predicate (e.g., eXtended WordNet (Harabagiu et al., 2003), Clark et al. (2008), Agerri & Peñas (2010)). None of these approaches have attempted to use the definitions to build ontological and lexical resources.

In contrast, our effort is focused on constructing rich ontological and lexical knowledge. We have produced a resource that has heretofore not been available: A broad-coverage verb lexicon in English, integrated with a substantial ontology.

We are applying the same techniques to extend COLLIE with nominals and properties (adjectives and adverbs). The

| Evaluation | % | Example ||||
|---|---|---|---|---|---|
| | | WordNet Sense | Definition | Derived Axiom | Comment |
| Completely Correct | 45.1 | abrade%2:35:01:: | "rub hard or scrub" | [AND [RUB-SCRAPE-WIPE ?ev ?agent ?affected] [INTENSE ?ev]] | Captures definition completely |
| Valid Entailment | 17.3 | agitate%2:38:01:: | "move or cause to move back and forth" | [OR [CAUSE-MOVE ?ev ?agent ?affected] [CAUSE-EFFECT ?agent ?affected [MOVE-BACK-AND-FORTH ?affected]] | True but first disjunct not complete due to wrong scoping of disjunction |
| Incorrect, undetected | 20.9 | catapult%2:35:00:: | "to shoot forth or …" | [OR [EVOKE-INJURY ?agent ?affected] … | Parser picked wrong sense of "shoot" |
| Rejected | 16.7 | ask%2:32:05:: | "consider obligatory" | [BELIEVE ?experiencer [NECESSARY ?formal]] | The definition indicates a stative, inconsistent with TRIPS-WordNet mapping |

Table 10: Evaluation of axioms, with (abbreviated) examples

```
ONT::PINION-WN23500
Parent: ONT::CONFINE
Arguments:
    AGENT {PHYS-OBJ ORIGIN=NATURAL}
    AFFECTED {PHYS-OBJ ORIGIN=NATURAL}
Definition:
[AND [ATTACH :agent ?agent :affected ?x]
    [EXTERNAL-BODY-PART :id ?x :figure ?affected]]
```

Figure 11: Derived concept for one sense of pinion

```
Pinion
    LF-PARENT  ONT::PINION-WN23500
    TEMPL  AGENT-AFFECTED-XP-TEMPL

Shackle
    LF-PARENT  ONT::PINION-WN23500
    TEMPL  AGENT-AFFECTED-XP-TEMPL
```

Figure 12: Lexical entries derived from *pinion%2:35:00::*

```
EXPERIENCER-FORMAL-SUBJCONTROL-TEMPL
LSUBJ  (% NP (var ?x))  EXPERIENCER
LCOMP  (% CP (ctype s-to) (subj (var ?x)) FORMAL
```

Figure 13: The linking template for an EXPERIENCER subject control verb such as *want*

end result will provide the first broad-coverage deep semantic lexicon for English (or any other language for that matter).

## 6. Description of Resources

In this section we describe the overall structure of COLLIE. The resource consists of two core databases, one for the ontology and one for the lexicon. Although we focused our discussion on the verb component of COLLIE, since COLLIE-V is built by augmenting the TRIPS ontology and lexicon, the resource also includes entries from TRIPS for other parts-of-speech in the same representation. We will use the same approach described here to expand COLLIE for other parts-of-speech in a later release.

COLLIE is available in both XML and JSON formats, and the JSON files are accompanied by a Python library that reads and reasons over them. The resource and more detailed documentation are available for download at

https://tripslab.github.io

At this site we also provide several supplementary resources, including the original TRIPS ontology and lexicon, and a hybrid ontology in which the TRIPS ontology is augmented with some of the information from the TRIPS-WordNet mappings.

In addition to the ontology and lexicon, one can also access the codes of several versions of the TRIPS parser that can use this data, as well as a number of supporting tools.

The code and data files can be downloaded and installed locally. There are also web versions (with associated APIs) for browsing COLLIE and other versions of the ontology and lexicon as well as parsing text online using these resources.

### 6.1 The Extended Ontology

COLLIE-V contains 6516 concepts, 5536 of which were derived automatically. Each concept entry identifies its parent concept, the set of arguments (semantic roles) and selectional preferences on those roles, and the entailment axioms. For instance, the concept ONT::PINION-WN23500 (see Figure 11) was created as a new subtype of the existing TRIPS type ONT::CONFINE using the techniques described in this paper. This was derived from the WordNet entry *pinion%2:35:00::* (bind the arms of). The specification identifies two core arguments, both with preferences for physical objects of natural origin. The definition indicates that pinioning something involves attaching some of its body parts. Note the current TRIPS ontology is not fine-grained enough to distinguish between *arm* and other external body parts. When the ontology is extended with noun concepts using the same techniques, we expect there would be a derived concept for *arm*.

### 6.2 The Extended Lexicon

COLLIE-V contains lexical entries for 9654 verbs, 7087 of which were derived automatically. Note this is less than the total verbs in WordNet because some verbs were not successfully processed. Many of the verbs in COLLIE have multiple entries as they have multiple senses in the ontology. As an example of some new entries, processing the synset containing the sense *pinion%2:35:00::* generated two new lexical entries, *pinion* and *shackle*, as shown in Figure 12, since the synset containing *pinion%2:35:00::* also contains *shackle%2:35:01::* Each entry identifies the ontology type ONT::PINION-WN23500 the lexical item belongs to (and thus the semantic roles, selectional preferences and entailment axioms) and one or more linking templates. In this case we have the simple transitive where the subject is AGENT and the direct object is AFFECTED. Not shown in Figure 12 or our totals are the derived forms for the tense and number variations, in this case *pinions* (3$^{rd}$ person present), *pinioned* (past and past participle), and *pinioning* (present participle).

### 6.3 Templates

The linking templates connect the semantic representations to their possible realizations in language. Given our method, no new linking templates can be derived from reading the definitions, but we have not yet come across a verb whose behavior is not covered by the set of existing templates for the verbs in the TRIPS lexicon. The linking templates resemble the same constructs in VerbNet but are specified in terms of the grammatical relations between the verb and the arguments rather than their positions in the sentence. This allows us to provide a uniform analysis for constructions involving the passive use and other movement phenomena.

The templates also allow detailed specification of interrelations between the arguments. For example, with subject control verbs such as *want*, as in *I want to sing*, the AGENT of the singing event is the EXPERIENCER of the wanting event. Figure 13 shows the linking template for such subject control verbs. The template states that the logical subject (LSUBJ) maps to the EXPERIENCER role and is a noun phase with identifier ?x, whereas the logical complement (LCOMP) maps to the FORMAL role and is a to-infinitive clause with the subject being the same ?x. In other words, the subject (EXPERIENCER) of *want* is also the subject (AGENT) of *sing*.

### 6.4 The Python Library

The `pytrips` library provides a native python interface to the TRIPS ontology and lexicon. The lexicon provides access to morphological and syntax features as well as syntactic templates. The ontology provides access to semantic roles and features, as well as mappings to WordNet. The library implements simple type subsumption, mapping lemmas to candidate types, and contains basic functions for navigating the ontology. Similarity metrics, including Wu-Palmer, are also provided. `pytrips` is available on `pypi`. The necessary data files are available as a separate accompanying package, `jsontrips`.

## 7. Acknowledgements

This research has been supported by DARPA ARO contract W911NF-15-1-0542 and the Office of Naval Research N00014-19-1-2308.

## 8. Bibliographical References